%% file: main.tex
\documentclass[10pt,twocolumn,letterpaper]{article}

\usepackage{wacv}              

\definecolor{wacvblue}{rgb}{0.21,0.49,0.74}
\usepackage[pagebackref,breaklinks,colorlinks,allcolors=wacvblue]{hyperref}
\usepackage{graphicx}
\usepackage{multirow} 
\usepackage{subcaption}
\usepackage{algorithm}
\usepackage{algpseudocode}
\usepackage{afterpage}
\usepackage{textgreek}
\usepackage{booktabs}
\usepackage{amsthm}
\usepackage{enumitem} 
\usepackage{verbatim}
\usepackage{amssymb}
\usepackage{wrapfig}
\usepackage{listings}
\usepackage[accsupp]{axessibility}  

\makeatletter
\DeclareRobustCommand\onedot{\futurelet\@let@token\@onedot}
\def\@onedot{\ifx\@let@token.\else.\null\fi\xspace}

\def\eg{\emph{e.g}\onedot}

\def\etc{\emph{etc}\onedot}

\newcommand{\reffig}[1]{Fig.~\ref{#1}}

\def\methodName{\textit{MoDOT}}

\def\syntheticBenchmarkName{\textit{OB-Hypersim}}

\def\lossName{\textit{OBDCL}}

\def\moduleName{\textit{CASM}}

\newcommand{\xlt}[1]{#1}


\def\wacvPaperID{1564} 
\def\confName{WACV}
\def\confYear{2026}

\title{Occlusion Boundary and Depth: Mutual Enhancement via Multi-Task Learning}

\newcommand\blfootnote[1]{%
  \begingroup
  \renewcommand\thefootnote{}\footnote{#1}%
  \addtocounter{footnote}{-1}%
  \endgroup
}

\author{
    Lintao XU\textsuperscript{1}
    \and
    Yinghao WANG\textsuperscript{2}
    \and
    Chaohui WANG\textsuperscript{1}$^{\dagger}$
    \and
    \and 
    {\textsuperscript{1} \small LIGM, Univ Gustave Eiffel, École des Ponts, CNRS, France } 
    \and
    {\textsuperscript{2} \small INFRES, Télécom Paris, Institute Polytechnique de Paris, France}
}

\begin{document}
\maketitle

\input{sections_round2/0_abstract}
\input{sections_round2/1_introduction}
\input{sections_round2/2_related_work}
\input{sections_round2/3_method}

\input{sections_round2/4_experiment}

\input{sections_round2/5_conclusion}



{
    \small
    \bibliographystyle{ieeenat_fullname}
    \bibliography{main}
}

\end{document}

%% file: sections_round2/0_abstract.tex
\begin{abstract}

Occlusion Boundary Estimation (OBE) identifies boundaries arising from both inter-object occlusions and self-occlusion within individual objects.
This task is closely related to Monocular Depth Estimation (MDE), which infers depth from a single image, as Occlusion Boundaries (OBs) provide critical geometric cues for resolving depth ambiguities, while depth can conversely refine occlusion reasoning.
In this paper, we aim to systematically model and exploit this mutually beneficial relationship. To this end, we propose \methodName, a novel framework for joint estimation of depth and OBs, which incorporates a new Cross-Attention Strip Module (\moduleName) to leverage mid-level OB features for depth prediction, and a novel OB-Depth Constraint Loss (\lossName) to enforce geometric consistency.
To facilitate this study, we contribute~\syntheticBenchmarkName, a large-scale photorealistic dataset with precise depth and self-occlusion-handled OB annotations.
Extensive experiments on two synthetic datasets and NYUD-v2 demonstrate that \methodName~achieves significantly better performance than single-task baselines and multi-task competitors. Furthermore, models trained solely on our synthetic data demonstrate strong generalization to real-world scenes without fine-tuning, producing depth maps with sharper boundaries and improved geometric fidelity. Collectively, these results underscore the significant benefits of jointly modeling OBs and depth. \xlt{Code and resources are available at \href{https://github.com/xul-ops/MoDOT}{HERE}.}

\end{abstract}

%% file: sections_round2/1_introduction.tex
\section{Introduction}
\label{sec:introduction}

\begin{figure*}
    \centering
    \includegraphics[width=0.98\linewidth]{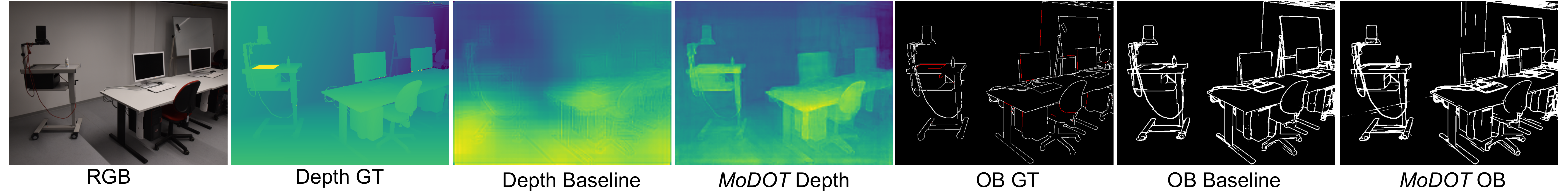}
    \vspace{-9pt}
    \caption{\textbf{Comparison of \xlt{zero-shot} results on real-world example \xlt{from iBims-1~\cite{koch2018ibims1}}.} Significant self-occlusion boundaries are highlighted in red. The proposed multi-task framework, \methodName, is jointly trained to predict both depth and OBs. Compared to the single-task depth baseline~\cite{yuan2022newcrfs}, which is trained solely with depth supervision, \methodName~produces more accurate and structurally consistent depth maps on unseen scenes, particularly around OBs, \xlt{while adding only 11M parameters (4\% of the depth baseline)}. Compared to the single-task OB baseline, \methodName~tends to predict more OBs (resulting in higher recall) but introduces additional noise, leading to a slight drop in F-score. All methods are trained exclusively on our proposed synthetic dataset.}
    \label{fig:teaser}
    \vspace{-9pt}
\end{figure*}

Occlusions, arising from overlapping 3D surfaces projections, are ubiquitous in 2D images of natural scenes, as illustrated in~\reffig{fig:teaser}. 
\blfootnote{\hspace{-15pt}$^{\dagger}$ \xlt{Corresponding author}\\ }
OBs delineate regions where one object occludes another or where self-occlusion occurs within a single object, providing pixel-level geometric cues that are critical for 3D scene understanding.
This characteristic allows OBs to convey finer structural details than conventional object contours, clearly distinguishing them from ordinary edges and semantic boundaries.
OBs inherently encode depth discontinuities and surface orientation changes, making them strongly correlated with depth estimation—a connection that remains underexplored in prior work.

Monocular depth estimation remains an ill-posed problem due to the inherent ambiguity of recovering 3D structure from a single 2D image.
Recent methods (\eg,~\cite{yuan2022newcrfs,shao2023nddepth,bhat2021adabins,ke2024repurposing,patni2024ecodepth,yang2024depthanything,hariat2025improved}) have advanced depth estimation through architectural innovations, improved prediction strategies, and by leveraging large-scale unlabeled data.
However, accurately resolving fine-grained depth boundaries and handling occlusions remain challenging, as depth ambiguities are most pronounced in these regions.

Multi-task (MT) learning offers a promising learning scheme for jointly optimizing related tasks by leveraging their inherent correlations. It has proven effective for dense prediction problems (\eg, depth, segmentation), as learning multiple related tasks simultaneously often leads to significant performance gains~\cite{ye2023taskprompter,ye2022invpt}. 
For depth and OBs, their mutual relationship—occlusions imply depth discontinuities, and abrupt depth changes signal occlusions—creates a natural synergy: OB priors can disambiguate depth near boundaries, while depth priors can refine OB localization.

Despite their interdependence, joint learning of depth and OBs remains underexplored—primarily because OBs that ignore self-occlusions often coincide with object contours, which are rarely studied as a standalone topic. Consequently, their bidirectional geometric constraints are seldom modeled within a unified framework. Furthermore, the lack of real-world datasets with precise, pixel-level correspondence between OBs and depth poses a significant limitation.
Recently, the availability of OB-FUTURE~\cite{xu2024iobe}—a synthetic dataset with self-occlusion-handled OB annotations—and our proposed photorealistic dataset~\syntheticBenchmarkName, where OBs include self-occlusion Ground-Truth (GT) labeled using the OB generation method from P2ORM~\cite{qiu2020P2ORM}, has made it feasible to begin bridging the gap between OBE and MDE using synthetic data.
Leveraging these resources, we propose the first integrative multi-task framework that jointly learns both tasks by exploiting their bidirectional geometric constraints through architectural innovation, boundary-aware optimization, and synthetic data engineering.

To reconcile the conflicting feature learning objectives of continuous depth regression and discrete boundary estimation, we design a Cross-Attention Strip Module (\moduleName) that dynamically fuses OB guidance with depth features. 
Meanwhile, multi-scale strip convolution aligns anisotropic receptive fields with the long and narrow structure of OBs, enabling the capture of fine-grained depth discontinuities for improved depth prediction. In addition, our devised Enhanced Image Path (\emph{EIP}) injects high-frequency, low-level edge cues into both OB and depth prediction refinement via an auxiliary image encoding module. To orchestrate the deep network components, we devise a tripartite loss function that unifies scale-invariant logarithmic regularization for global depth coherence, class-balanced cross-entropy for handling extreme foreground-background imbalance in OB estimation, and a novel OB-Depth Constraint Loss (\lossName) that explicitly enforces depth divergence across boundary pixels through adaptive margin constraints—mathematically codifying the mutual reinforcement between the two tasks. 

In a nutshell, our main contributions can be summarized as the following three aspects:

\begin{itemize}
    \item We systematically model and validate the mutually beneficial relationship between discrete structural prediction (OBE) and continuous geometric regression (MDE) through extensive experiments. For this purpose, we propose \methodName, the first \textbf{M}ulti-task \textbf{O}cclusion and \textbf{D}epth \textbf{O}ptimization \textbf{T}raining framework, which simultaneously estimates monocular depth and OBs and notably outperforms both single-task and multi-task competitors.
   
    \vspace{3pt}
    \item We introduce the novel \lossName~loss and the simple yet effective \moduleName~module, both of which demonstrate strong performance in our experiments. Together, they systematically capture the mutual geometric correlation between long, narrow mid-level occlusion boundaries and high-amplitude depth discontinuities. 
  
    \vspace{3pt}
    \item We construct a photorealistic dataset, \syntheticBenchmarkName, sourced from Hypersim~\cite{roberts2021hypersim}, which provides pixel-perfect annotations for both depth maps and self-occlusion-aware OBs across diverse indoor scenes. Models trained on this dataset demonstrate strong generalization to real-world scenes from the depth datasets iBims-1~\cite{koch2018ibims1} and DIODE~\cite{vasiljevic2019diode}, without requiring domain adaptation or fine-tuning, as illustrated in~\reffig{fig:teaser} and~\reffig{fig:transfer}.
\end{itemize}

%% file: sections_round2/2_related_work.tex
\section{Related Work}
\label{sec:related_work}

\par\noindent\textbf{Edge- and Occlusion-Aware Depth Estimation.}
MDE has been extensively studied in a series of works (\eg,~\cite{bhat2021adabins,ke2024repurposing,agarwal2023pixelformer,liu2023va,patni2024ecodepth}) and often relies on auxiliary geometric cues to constrain solution spaces. Existing approaches typically incorporate surface normals~\cite{shao2023nddepth,long2021adaptive} or semantic segmentation~\cite{zhang2019pad,mousavian2016joint} as supplementary priors. Recent advances particularly emphasize the critical role of edge and/or occlusion cues in MDE. For instance, 
the twin-surface representation in~\cite{imran2021depth} employed an asymmetric loss to refine depth predictions near OBs, while~\cite{ramamonjisoa2019sharpnet} introduced a joint learning approach of object contours and depth for mutual enhancement. \cite{talker2024mind} proposed learning depth edge detection from synthetic data and using it to supervise MDE. \cite{hariat2025improved} improved self-supervised MDE by leveraging pre-semantic contours to resolve ambiguities in low-texture regions. \xlt{\cite{gui2025depthfm, ke2024repurposing, Pham_2025_CVPR} leveraged diffusion priors to enhance the sharpness of boundaries. \cite{xu2025blurry} proposed a blurry-edges representation for a robust, edge-aware depth-from-defocus method in photon-limited images. \cite{bochkovskii2025depthpro} built a foundation model for zero-shot metric MDE, achieving sharp boundary delineation and high-frequency details.} While these methods demonstrate the value of occlusion reasoning, current frameworks predominantly focus on \textit{object-level} occlusion contours (\eg, boundaries between distinct objects), the more subtle \textit{self-occlusion} remains underexplored. This oversight limits these frameworks' ability to resolve more accurate depth ambiguities in fine-grained structures.

\vspace{3pt}
\par\noindent\textbf{Occlusion Boundary Estimation with Depth.}
OBE has been extensively studied as both an independent task~\cite{xu2024iobe,wang2020occlusion,he2010occlusion} and through occluder-occludee relationships~\cite{wang2016doc,wang2019doobnet}. The intrinsic connection between OBs and depth stems from OBs' encoding of depth discontinuities~\cite{teo2015fast,qiu2020P2ORM,hambarde2024occlusion}. Early works used occlusion cues for depth ordering~\cite{feldman2008motion} or pseudo-depth maps for OB recovery~\cite{he2010occlusion}. 
Recent advances extended OB estimation to self-occlusions, with \cite{wang2020occlusion} formulating a geometric OB definition distinct from segmentation contours. For instance, P2ORM~\cite{qiu2020P2ORM} established a 2D OB annotation protocol and demonstrated that self-occlusion-aware OBs can effectively improve depth inference. 
While some works incorporated OBs to enhance depth estimation~\cite{qiu2020P2ORM,hambarde2024occlusion,Zhang_2018_CVPR}, the joint modeling of self-occlusion-aware OBs and depth remains underexplored due to the inherent challenges mentioned above.

\vspace{3pt}
\par\noindent\textbf{MT Learning for Dense Predictions}
often couples depth estimation with surface normal prediction and/or semantic segmentation, leveraging task correlations to enhance overall performance. Recently, several multi-task frameworks \xlt{(\eg,~\cite{bruggemann2021atrc,vandenhende2020mtinet,ye2022invpt,ye2023taskprompter,jiang2024mlore,yangmulti2025difmt})} have been proposed to jointly optimize depth estimation, surface normal estimation, semantic segmentation, and object boundary detection in a unified model. Based on real-world datasets like NYUD-v2~\cite{silberman2012nyudv2}, these methods demonstrated that jointly optimizing multiple objectives can lead to performance improvements across all four tasks.
However, while object boundaries are commonly used as auxiliary supervision, they are generally less informative than OBs, as object boundaries represent only a subset of OBs. More importantly, the relationship between edge/boundary cues and depth has not been explicitly modeled in existing MT methods. In our method, we address this overlooked connection by explicitly exploring the synergy between depth and OBs.
To the best of our knowledge, we propose the first multi-task learning framework specifically designed for joint learning of OBs and depth.

\vspace{3pt}
\par\noindent\textbf{Occlusion Boundary-Depth Dataset.} 
There are very few datasets that provide both pixel-level self-occlusion-handled OBs and depth annotations.
As a result, prior works~\cite{xu2024iobe,qiu2020P2ORM} have developed 2D occlusion-simulation-based and 3D occlusion-definition-based approaches to automatically generate large-scale synthetic datasets for depth refinement and interactive benchmark construction.

\begin{figure*}[ht!]
  \centering
  \small
  \begin{minipage}{0.88\textwidth}
    \centering
    \includegraphics[width=\textwidth]{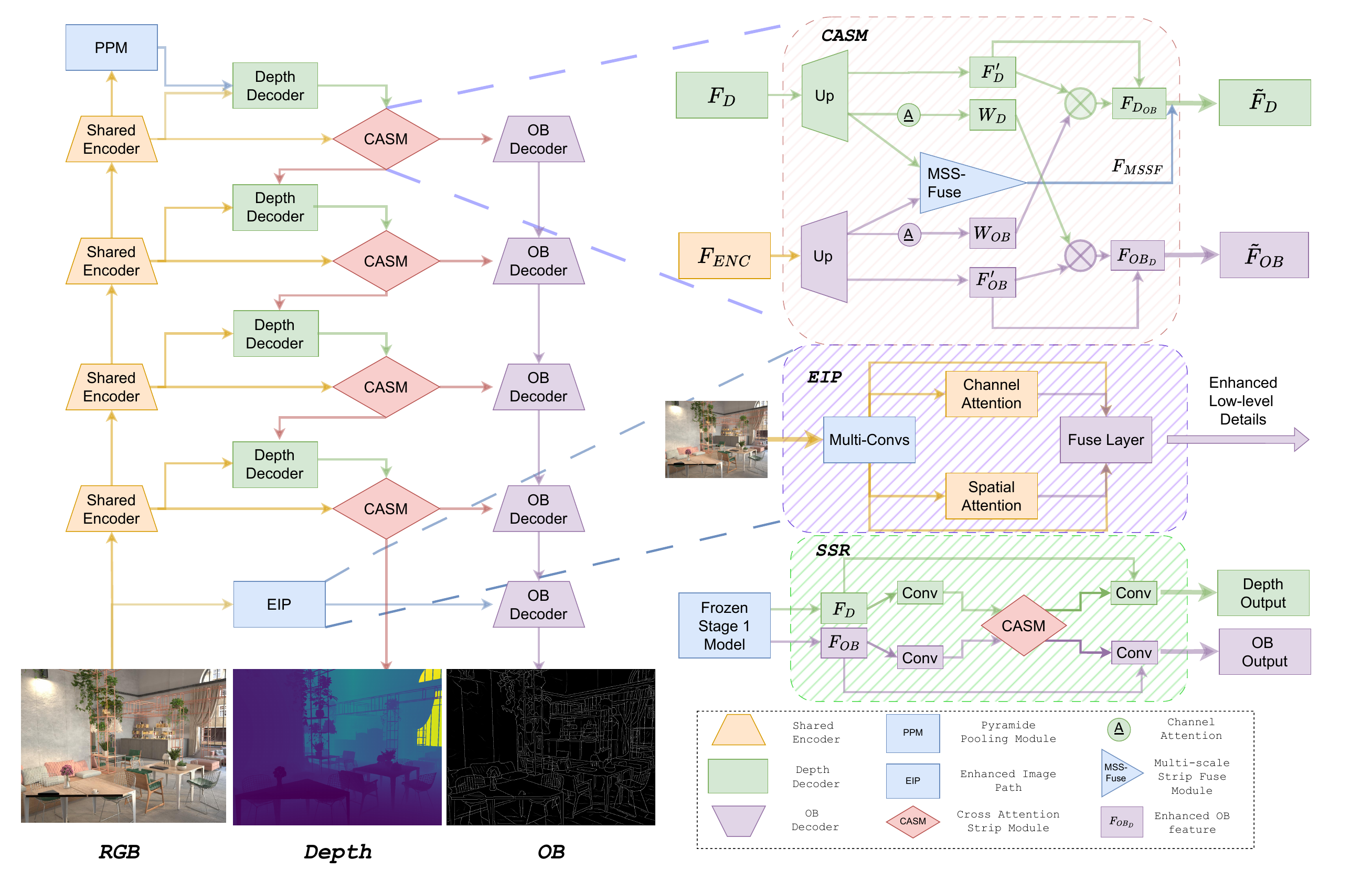} 
  \end{minipage}
  \vspace{-9pt}
  \caption{\textbf{Overall framework of~\methodName.} On the left is the main model structure of \methodName~used in the first training stage; on the right are the detailed structures of~\moduleName, \emph{EIP}, and Second Stage Refinement (\emph{SSR}). Our method uses a shared encoder that extracts hierarchical image features, which are then separately fed into the depth decoder and the OB decoder. The designed \moduleName~facilitates communication between the encoder and the two decoders, enhancing feature interaction for both depth and OB predictions. A PPM aggregates global high-level context for the first-stage depth decoder, while the proposed \emph{EIP} captures enhanced local low-level image features for the final OB decoder and the subsequent \emph{SSR}. \emph{SSR} is incorporated to further refine depth and OB features on full-resolution images.}
  \label{fig.model_pipeline}
  \vspace{-10pt}
\end{figure*}

%% file: sections_round2/3_method.tex
\section{Method}
\label{sec:method}
\vspace{-3pt}

\textbf{Overall pipeline} of the proposed \methodName~is illustrated in \reffig{fig.model_pipeline}, which adopts a two-stage architecture:
    \textbf{Stage One} (left side of \reffig{fig.model_pipeline}) serves as the core component, trained on cropped images. It consists of a shared encoder that hierarchically encodes input images into shared features for two task-specific decoders: the OB decoder and the Depth decoder. The proposed \moduleName~facilitates communication between the encoder and decoders. The \emph{EIP} and Pyramid Pooling Module (PPM) further enhance the OB and depth predictions. The output features from Stage One are upsampled to generate predictions and simultaneously forwarded to the Second Stage Refinement (\emph{SSR}) for further enhancement.
    \textbf{Stage Two} (\emph{SSR}, shown in~\reffig{fig.model_pipeline}) 
    refines both the depth features (extracted from the final \moduleName~of Stage One) and OB features (from the last OB decoder in Stage One). This refinement process incorporates the full-resolution input image and is primarily implemented through an additional \moduleName.
    
In the following sections, we describe the proposed \moduleName, the overall loss formulation including our loss constraint \lossName, and the remaining network components.

\subsection{Cross-Attention Strip Module~(\textbf{\moduleName})}

To bridge the gap between continuous depth prediction and discrete OB estimation, we design the simple and effective~\moduleName~to facilitate feature communication and enhancement during Stage one/two. The detailed architecture of \moduleName~is shown in the top-right of~\reffig{fig.model_pipeline}.

In Stage One, features from the shared encoder ($F_{ENC}$) branch into two paths: (1) depth features $F_{D}$ obtained by passing $F_{ENC}$ through the depth decoder, and (2) direct OB features, where $F_{ENC}$ is fed into the \moduleName~and then into the OB decoder. In Stage Two, the inputs to the \moduleName~($F_{D}$ and $F_{OB}$) are taken from the Stage One as described above.

Given a pair of depth and OB features, we first upsample both feature maps and align their channel dimensions facilitate effective fusion. The resulting upsampled and channel-aligned features ($F'_{D}$ and $F'_{OB}$) are then passed through two parallel branches:

\begin{itemize}
    \item \textbf{Cross-Channel Attention}: Dual attention branches operate in parallel to generate channel-wise attention weights for depth ($F'_{D}$) and OB features ($F'_{OB}$), respectively. These modality-specific weights are cross-applied through element-wise multiplication: depth-derived attention weights $W_D$ recalibrate OB feature channels while OB-generated weights $W_{OB}$ re-weight depth feature channels. Channel attention~\cite{zhang2018image} effectively emphasizes task-relevant features in vision tasks. Building on this, our cross-attention-based reciprocal re-weighting mechanism adaptively enhances the depth and OB features to obtain boundary-salient depth features $F_{D_{OB}}$ and depth-aware boundary features $F_{OB_{D}}$, leveraging the geometry complementary relationship between the two tasks through channel-wise recalibration.

    \item \textbf{Multi-Scale Strip Fuse Module (MSS-Fuse)}: The aligned features are concatenated and undergo hierarchical processing through a hybrid convolutional architecture that combines local context preservation with long-range dependency modeling. Specifically, we employ 1×7, 7×1, 1×11, and 11×1 strip convolutional layers, alongside standard 3×3 convolutional layers, to separately process the concatenated features. A final 1×1 convolutional layer is then applied to fuse these features and obtain the enhanced depth representation ($F_{MSSF}$).
    NeWCRFs~\cite{yuan2022newcrfs} improves depth estimation via window-based Conditional Random Field (CRF) decoders (adopted in~\methodName); however, they are inherently limited in capturing global context. To address this limitation and further leverage OB features to guide depth estimation, we incorporate multi-scale convolutional layers that inject global OB-aware information into the depth features. Furthermore, to effectively capture the elongated and narrow structures characteristic of OBs, we employ strip convolutional layers, whose effectiveness has been demonstrated in prior OB estimation work~\cite{feng2021mtorl,xu2024iobe}. These multi-scale strip convolutions expand horizontal and vertical receptive fields in alignment with OB geometry, thereby improving the integration of depth and OB features to enhance depth representation.

\end{itemize}

After passing through the two components, the depth features from MSS-Fuse ($F_{MSSF}$) and from the cross-channel attention branch ($F_{D_{OB}}$) are merged to form OB-enhanced depth representations $\tilde{F}_D$, which are then fed into the next depth decoder block. Meanwhile, the depth-enhanced OB features ($F_{OB_{D}}$ or $\tilde{F}_{OB}$) serve as input for the current OB decoder with side-output supervision. 
\moduleName~effectively facilitates communication between depth features and OB-aware encoder features, enhancing both modalities for improved decoding. 

\subsection{Overall Loss Functions and \textbf{\lossName}}

The tripartite loss $\mathcal{L}$ used for supervising the training consists of: (1) $\mathcal{L_{D}}$: the Scale-Invariant Logarithmic Loss (SILog)~\cite{eigen2014silog}, commonly used in monocular depth estimation~\cite{yuan2022newcrfs,bhat2021adabins,lee2019big}; (2) $\mathcal{L_{OB}}$: the Class-Balanced Cross-Entropy (CCE) Loss~\cite{feng2021mtorl}, used for OB estimation; and (3) $\mathcal{L_{C}}$: the proposed OB-Depth Constraint Loss (\lossName), which encourages depth discontinuities to align with ground-truth OB locations. The complete multi-task loss is formulated as:
\begin{equation}
    \mathcal{L} = w_{d} \cdot \mathcal{L_{D}} + w_{ob} \cdot \mathcal{L_{OB}} + w_{c} \cdot \mathcal{L_{C}}
\end{equation}
In the experiments, the corresponding loss weights for each loss are set to $w_{d}=1.2, w_{ob}=1.0, w_{c}=0.1$ for the SILog loss, CCE loss, and~\lossName, respectively. The detailed formulations of $\mathcal{L_{D}}$ and $\mathcal{L_{OB}}$ are provided in the Supplementary Materials (referred to as \emph{Supp.} hereafter).


Given a ground-truth OB map $B \in \{0,1\}^{H \times W}$ and a predicted depth map $D \in \mathbb{R}^{H \times W}$, we compute a depth difference map $\Delta \in \mathbb{R}^{H \times W}$. This is done by first generating shifted versions of $D$ ($D_\text{left}$, $D_\text{right}$, $D_\text{top}$, $D_\text{bottom}$) by displacing it by $n$ pixels ($n=1$ in our experiments) and then calculating the sum of absolute differences:
\begin{equation}
    \Delta = |D_\text{top} - D_\text{bottom}| + |D_\text{left} - D_\text{right}|.
\end{equation}
$\mathcal{L}_C$ is then defined as:
\begin{equation}
    \mathcal{L}_C = \frac{1}{\|B\|_1} 
    \sum_{(h,w)} B_{h,w} \cdot \big(1 - \Delta_{h,w}\big),
    \label{eq:obcdl}
\end{equation}
where $\|B\|_1$ denotes the $\ell_1$-norm of $B$, i.e., the number of OB pixels, which serves as a normalization constant.

The key idea behind the \lossName~is simple yet intuitive: we encourage the predicted depth values on the vertical and horizontal sides of a ground-truth OB to differ, reflecting the expected depth difference. This concept aligns with the mathematical definition in \cite{wang2020occlusion} and effectively leverages self-occlusion boundaries within objects, rather than relying solely on object contours. Meanwhile, ordinary edges—such as texture, illumination, or glass reflections—do not satisfy this constraint, further demonstrating that OBs are more suitable for joint learning with depth.

\begin{figure*}[ht!]
    \centering
    \includegraphics[width=0.93\linewidth]{images/result_v2.pdf}
    \vspace{-6pt}
    \caption{\textbf{Qualitative comparisons on the synthetic \syntheticBenchmarkName~and the real NYUD-v2 datasets.} Our \methodName~produces more accurate depth maps than the single-task depth baseline and achieves higher-recall OB predictions compared to the single-task OB baseline.}
    \vspace{-6pt}
    \label{fig:result}
\end{figure*}

\subsection{Remaining Network Components}
In this section, we detail the remaining components of the proposed~\methodName, including the shared encoder, the two task-specific decoders, our \emph{EIP}, and \emph{SSR}.

\vspace{3pt}
\par\noindent\textbf{Shared Encoder.}
We adopt the Swin Transformer~\cite{liu2021swin} as an encoder to hierarchically capture multi-scale image features. The encoder processes an input image of size $H \times W$ across four stages, progressively increasing patch resolutions from $4\times 4$ to $32\times 32$ pixels. Initial $4\times 4$ non-overlapping patches split the image into $\frac{H}{4}\times \frac{W}{4}$ tokens, followed by successive 2$\times$ downsampling in subsequent stages to generate $8\times 8$, $16\times 16$, and $32\times 32$ patch representations.
Within each stage, features are divided into fixed $N \times N$ windows, yielding$\frac{H_i}{N}\times \frac{W_i}{N}$ local regions at level $i$, where $H_i$ and $W_i$ denote the feature map dimensions.

\vspace{3pt}
\par\noindent\textbf{Depth Decoder and PPM.} 
We adopt the neural window-based fully connected CRFs depth decoder proposed in the strong baseline NeWCRFs~\cite{yuan2022newcrfs} as the decoder of our depth prediction branch. This decoder leverages a self-attention mechanism inspired by transformers~\cite{vaswani2017attention,dosovitskiy2020vit}, using query and key features to compute pairwise potentials—affinity scores between each CRF node (pixel) and all others within a local window.  Due to space constraints, we omit their detailed formulations of the energy function, unary potential, and pairwise potential. Additionally, similar to NeWCRFs, we incorporate a PPM~\cite{zhao2017ppm} to aggregate global contextual information. Specifically, global average pooling with scales of 1, 2, 3, and 6 is applied to the top-level encoder features to capture global context for the depth decoder. 

\vspace{3pt}
\par\noindent\textbf{Enhanced Image Path (\emph{EIP}).}
Prior work~\cite{feng2021mtorl,xu2024iobe} on OB estimation preserves critical low-level details by processing input images at native resolution through dedicated encoder modules. Building upon this foundation, we enhance their image path architecture through dual-attention integration—combining spatial and channel attention mechanisms. Initial low-level features extracted via convolutional layers undergo parallel enhancement through these complementary attention modules, followed by adaptive fusion and progressive decoding with skip connections from preceding OB decoder features to generate final boundary predictions.
\emph{EIP} not only provides low-level edge details for OB estimation in the first stage but also enhances depth prediction in SSR via more accurate OB cues.

\vspace{3pt}
\par\noindent\textbf{OB Decoder and Side-output Supervision.}
Our OB decoder employs a streamlined architecture—five sequential decoding blocks enabled by the EIP—avoiding complex designs. Each block first halves channels via $1 \times 1$ convolution, processes features through dual $3\times 3$ convolutions, then upsamples outputs for subsequent OB decoder blocks or additional side-output supervision. Following edge detection and OB estimation conventions, side-output supervision injects auxiliary gradients during training to enhance boundary precision, as validated in prior works~\cite{he2019bdcn,xu2024iobe,feng2021mtorl,soria2023dexined_ext}. 

\vspace{3pt}
\par\noindent\textbf{Second Stage Refinement (\emph{SSR}).} \emph{SSR} acts as an additional refinement module, trained with the first-stage model’s parameters frozen. In SSR, an additional \moduleName~and several simple convolutional layers are employed to further refine the final depth and OB features from the last \moduleName~and OB decoder of the first-stage model at full image resolution. These refined features are then passed to the final convolutional layers to produce the final OB/depth predictions.

%% file: sections_round2/4_experiment.tex
\section{Experiments}
\label{sec:experiment}

\begin{table*}[ht!]
\caption{Quantitative comparisons on two synthetic datasets. Our~\methodName~consistently outperforms MT baselines across both depth and OB estimation tasks. The best results are set in \textbf{bold}. Complete Depth and OB evaluation metrics for each dataset are provided in the \emph{Supp}.} 
  \label{tab-syn}
  \vspace{-6pt}
  \centering
  \small
  \scalebox{0.9}{
  \begin{tabular} {lcccc|cccc} 
      \toprule 
      \multicolumn{1}{c}{} & \multicolumn{4}{c}{\textbf{OB-FUTURE}} & \multicolumn{4}{c}{\textbf{\syntheticBenchmarkName}} \\
      \cmidrule(lr){2-5} \cmidrule(lr){6-9}
      \textbf{Method} & RMSE$\downarrow$  & Abs Rel$\downarrow$  & $\delta < 1.25\uparrow$ & OB-Recall$\uparrow$  & RMSE$\downarrow$  & Abs Rel$\downarrow$  & $\delta < 1.25\uparrow$ & OB-Recall$\uparrow$ \\
      \midrule
       Depth Baseline~\cite{yuan2022newcrfs} & 0.4524 & 0.1011 & 0.9215 & - & 0.6948 & 0.3123 & 0.4759 & -  \\
       OB Baseline (ours) & - & - & -  & 0.7655 & - & - & - & 0.8099 \\
      \midrule
       SharpNet~\cite{ramamonjisoa2019sharpnet} & 0.9535 & 0.1890 & 0.5966 & \textbf{0.9571} & 0.8551 & 0.4422 & 0.3561 & 0.7342  \\
       MTAN~\cite{liu2019mtan} & 0.5576 & 0.1218 & 0.8523 & 0.9238 & 0.8050 & 0.3765  & 0.4023 & 0.7430 \\
       PAD-Net~\cite{xu2018padnet} & 0.5447 & 0.1188 & 0.8627 & 0.9022  & 0.8404  & 0.4322  & 0.3866 & 0.6732 \\
       MTI-Net~\cite{vandenhende2020mtinet} & 0.5064 & 0.1106  & 0.8891 & 0.9125 & 0.7560 & 0.3746 & 0.4365 & 0.7490  \\
       InvPT~\cite{ye2022invpt} & 0.9335  & 0.2371  & 0.6122 & 0.3228 & 0.9018 & 0.5106 & 0.3555 & 0.8004 \\
       DenseMTL~\cite{lopes2023densemtl} & 0.5217 & 0.1106 & 0.8818 & 0.8927 & 0.7475 & 0.4095  & 0.4410 & 0.8520 \\
       \midrule
       Ours  & 0.3963  & 0.0901 & 0.9427 & 0.9090  & 0.6583 & 0.2963 & 0.5167 & 0.8670  \\
       Ours + \emph{SSR}  & \textbf{0.3809}  & \textbf{0.0843}  & \textbf{0.9518} & 0.9486 & \textbf{0.6537} & \textbf{0.2954}  & \textbf{0.5148} & \textbf{0.8732} \\
      \bottomrule 
  \end{tabular}
  }
\end{table*}

\begin{table*}[ht!]
  \caption{Quantitative comparisons on NYUD-v2 dataset. Our \methodName~already achieves superior results even without the additional \emph{SSR}.} 
  \label{tab-nyud}
  \centering
  \small
  \vspace{-6pt}
  \scalebox{0.9}{
  \begin{tabular} {lcccccc|cc} 
      \toprule 
      \textbf{Details} & RMSE$\downarrow$ & $RMSE_{log}\downarrow$  & Abs Rel$\downarrow$ & Sq Rel$\downarrow$  & log10$\downarrow$ & $\delta < 1.25\uparrow$ & OB-Recall$\uparrow$ & OB-Fscore$\uparrow$ \\
      \midrule
       Depth Baseline~\cite{yuan2022newcrfs} & 0.4619 & 0.1684 & 0.1363 & 0.0876 & 0.0558 & 0.8438 & - & - \\
       OB Baseline (ours)  & - & - & - & - & - & - & 0.5509 & 0.1736 \\            
       \midrule
       SharpNet~\cite{ramamonjisoa2019sharpnet} & 0.6188 & 0.2215 & 0.1888 & 0.1523 & 0.0777 & 0.7107 & 0.5072 & 0.1642 \\
       MTAN~\cite{liu2019mtan} & 0.5499 & 0.2043 & 0.1664 & 0.1241 & 0.0685 & 0.7655 & 0.5439 & 0.1544 \\ 
       PAD-Net~\cite{xu2018padnet} & 0.6404 & 0.2361 & 0.2009 & 0.1706 & 0.0810 & 0.6964 & 0.5681 & 0.1493 \\ 
       MTI-Net~\cite{vandenhende2020mtinet} & 0.5630 & 0.2004 & 0.1674 & 0.1303 & 0.0687 & 0.7650 & 0.5049 & 0.1596 \\ 
       InvPT~\cite{ye2022invpt} & 0.6574 & 0.2389 & 0.2087 & 0.1871 & 0.0830 & 0.6998 & 0.5746 & 0.1443 \\
       DenseMTL~\cite{lopes2023densemtl} & 0.5158 & 0.1852 & 0.1502 & 0.1081 & 0.0631 & 0.8006 & \textbf{0.6331} & 0.1616 \\ 
       TaskPrompter~\cite{ye2023taskprompter} & 0.4446 & 0.1548 & 0.1228 & 0.0758 & 0.0525 & 0.8591 & 0.2906 & 0.1626 \\
       MLoRE~\cite{jiang2024mlore} & 0.4576 & 0.1601 & 0.1285 & 0.0819 & 0.0544 & 0.8493 & 0.5059 & 0.1616 \\    
       \midrule
       Ours & 0.4174 & 0.1475 & 0.1169 & 0.0692 & 0.0498 & 0.8741 & 0.6287 & 0.1729 \\ 
       Ours + \emph{SSR} & \textbf{0.4137} & \textbf{0.1460} & \textbf{0.1155} & \textbf{0.0679} & \textbf{0.0492} & \textbf{0.8767} & 0.6244 & \textbf{0.1863} \\ 
      \bottomrule 
  \end{tabular}
  }
  \vspace{-6pt}
\end{table*}

In this section, we present further details about the experimental setup, quantitative and qualitative results, and model ablation analysis~\footnote{More qualitative \& quantitative results and implementation details are provided in the Supplementary Material (\emph{Supp}).}.

\par\noindent\textbf{Datasets.}
The experiments are mainly conducted on the following synthetic and real-world datasets:

\textbf{OB-FUTURE} is built using the 3D occlusion boundary generation method proposed in~\cite{xu2024iobe}, applied to 3D indoor scenes from the \emph{3D-FUTURE} dataset~\cite{fu20213dfuture}. This OB generation process is directly derived from the mathematical definition in~\cite{wang2020occlusion}, enabling more accurate OB annotations that include complete self-occlusions. The OB-FUTURE dataset consists of 17,267 training images and 1,869 test images, each with a resolution of $1080 \times 1080$.

\textbf{NYUD-v2} dataset~\cite{silberman2012nyudv2} for multi-task learning consists of 795 training images and 654 testing images with a valid resolution of $560 \times 425$, covering diverse indoor scenes such as offices, living rooms, \etc. The dataset provides dense annotations including monocular depth and object boundaries/contours. As object boundaries constitute a subset of OBs, these annotations effectively offer partial OB labeling.

\textbf{\syntheticBenchmarkName} is constructed using the 2D occlusion boundary generation method from P2ORM~\cite{qiu2020P2ORM}, applied to the source data from Hypersim~\cite{roberts2021hypersim}. Since P2ORM is unable to generate complete object occlusion contours, we supplemented its output by producing instance segmentation maps to extract additional boundaries. These were combined with P2ORM-generated boundaries to form pseudo-GT OB annotations. Due to P2ORM’s sensitivity to depth noise in distant regions, we tested multiple parameter settings for its generation process and manually removed scenes containing noisy OB labels and inappropriate/broken scenes. The final resulting dataset contains 27,536 training images and 3,311 testing images, all at a resolution of $1024 \times 768$. Additional details on dataset generation, filtering, and visualization are provided in the \emph{Supp}.

\vspace{2pt}
\par\noindent\textbf{Baselines.} 
Two single-task baselines were built and evaluated to quantify the improvements from MT learning: the depth baseline adopted NeWCRFs decoders~\cite{yuan2022newcrfs}, while the OB baseline utilized our proposed OB decoder, both single-task baselines used the encoder identical to \methodName. For MT baselines' comparisons, we evaluated CNN-based approaches~\cite{ramamonjisoa2019sharpnet, liu2019mtan, xu2018padnet, lopes2023densemtl, vandenhende2020mtinet}, along with transformer-based methods~\cite{ye2022invpt, ye2023taskprompter,jiang2024mlore}. As these MT baselines lack official performance metrics for simultaneous dual-task evaluation, we rigorously retrained them under identical protocols including training strategies, data augmentations, \etc 
Notably, transformer-based methods~\cite{ye2023taskprompter, jiang2024mlore} built on ViT-L backbones rely on fixed positional embeddings and rigid feature map shapes tailored to standard dataset resolutions. These constraints lead to significant performance degradation on our synthetic datasets, and simple resizing does not resolve the issue. Consequently, we report their results only on the real-world NYUD-v2 benchmark.

\vspace{2pt}
\par\noindent\textbf{Evaluation metrics.}
For depth, we adopted standard evaluation metrics commonly used in prior work: Root Mean Squared Error (RMSE), RMSE in log space ($RMSE_{log}$), Absolute Relative Error (Abs Rel), Squared Relative Error (Sq Rel), Logarithmic Error (log10), and Accuracy under threshold ($\delta_1$). The evaluation mode in NewCRFs~\cite{yuan2022newcrfs} were used here. For OBs, due to the high computational cost and long runtime of traditional MATLAB-based evaluation on large, high-resolution images~\cite{xu2024iobe,qiu2020P2ORM,feng2021mtorl}, we instead used a simplified and efficient approach. Specifically, we computed the recall (OB-Recall) and F-score (OB-Fscore) by comparing the estimated edges under a fixed probability thresholding of $0.7$ against the ground truth OBs. 

\begin{figure*}[ht!]
    \centering
    \includegraphics[width=0.88\linewidth]{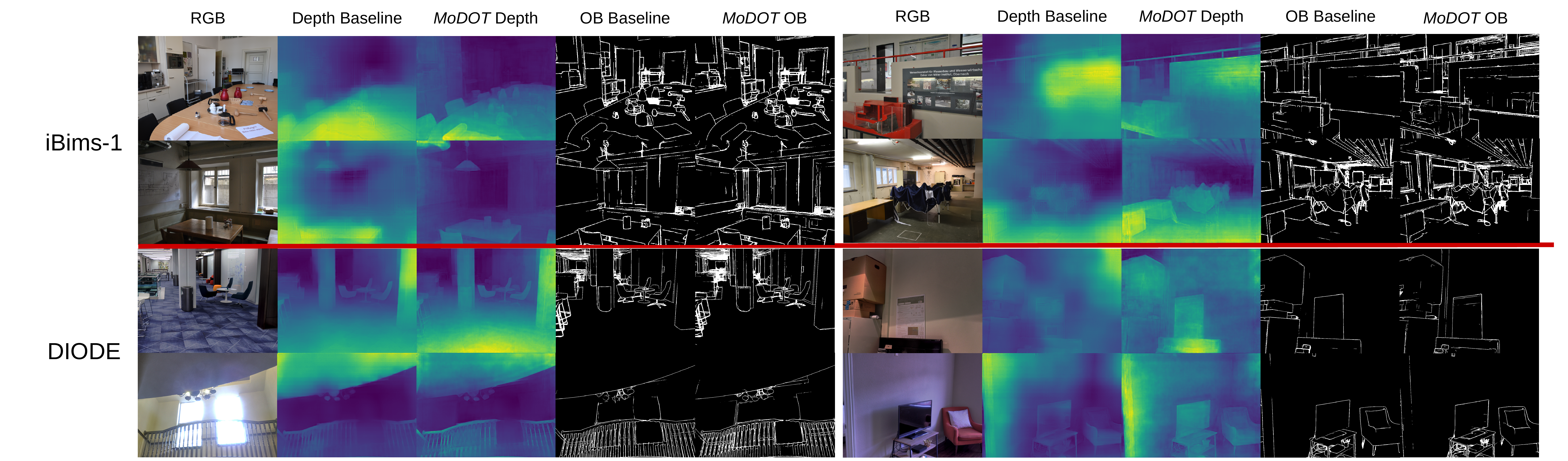}
    \vspace{-6pt}
    \caption{\xlt{Zero-shot} comparison on real scenes from the iBims-1~\cite{koch2018ibims1} and DIODE~\cite{vasiljevic2019diode} datasets, where the single-task baselines and our~\methodName~are trained exclusively on proposed synthetic~\syntheticBenchmarkName.
    }
    \label{fig:transfer}
    \vspace{-6pt}
\end{figure*}

\subsection{Qualitative \& Quantitative Results}

In-domain quantitative visualization comparisons with two single-task baselines are shown in~\reffig{fig:result}.
Further in~\reffig{fig:transfer}, we presented additional cross-domain results of \methodName~on the real-world datasets iBims-1~\cite{koch2018ibims1} and DIODE~\cite{vasiljevic2019diode}, where the models were trained solely on the synthetic~\syntheticBenchmarkName~dataset without any domain adaptation or fine-tuning. Notably, compared to the single-task depth baseline, our multi-task model with additional OB supervision significantly improved depth estimation quality, especially in geometric structure accuracy.

The quantitative comparisons with the MT baselines on the two synthetic datasets are presented in Table~\ref{tab-syn}. Notably, even without additional refinement stages, the proposed \methodName~already significantly outperformed the compared methods, highlighting the effectiveness of our design. Specifically, on OB-FUTURE, our method outperformed the best MT baseline in terms of RMSE, Abs Rel, and OB Recall. Similarly, on \syntheticBenchmarkName, our approach achieved superior performance across the same metrics. 

To evaluate the performance of our method on real-world data, we further report results on the widely used NYUD-v2 dataset. We evaluated both depth estimation and object boundary detection (which, as discussed, constitutes a subset of OBs). The results summarized in Table~\ref{tab-nyud}, further validate the effectiveness and advantages of \methodName.

\subsection{Method Ablation Analysis}

\begin{table*}[ht]
  \caption{Performance on~\syntheticBenchmarkName~when jointly training depth with different auxiliary ground truths in place of OBs. The results highlight the effectiveness of self-occlusion-aware OB supervision in improving depth estimation.} 
  \vspace{-6pt}
  \label{tab-AGT}
  \centering
  \small
  \scalebox{0.99}{
  \begin{tabular} {lcccccc} 
      \toprule 
      \textbf{Auxiliary GT} & RMSE$\downarrow$ & $RMSE_{log}\downarrow$  & Abs Rel$\downarrow$ & Sq Rel$\downarrow$  & log10$\downarrow$ & $\delta < 1.25\uparrow$ \\
       \midrule
       None & 0.6948 & 0.3739 & 0.3123 & 0.2481 & 0.1342 & 0.4759 \\
       \midrule
       Occlusion boundary & \textbf{0.6583} & \textbf{0.3463} & 0.2963 & \textbf{0.2279} & \textbf{0.1235} & \textbf{0.5167} \\
       Instance contour & 0.6736 & 0.3580 & 0.3195 & 0.2496 & 0.1278 & 0.4991 \\
       Semantic contour & 0.6798 & 0.3613 & 0.3131 & 0.2453 & 0.1294 & 0.5013 \\
  
       Edge  & 0.6786 & 0.3557 & 0.2988 & 0.2409 & 0.1273 & 0.5082 \\ 
       
       \midrule
       Semantic segmentation & 0.6699 & 0.3591 & \textbf{0.2883} & 0.2315 & 0.1281 & 0.5058 \\
       Surface normal & 0.6729 & 0.3603 & 0.3025 & 0.2377 & 0.1275 & 0.5108 \\  
      \bottomrule 
  \end{tabular}
  }
  \vspace{-6pt}
\end{table*}

\begin{table*}[ht!]
  \caption{Ablation results on OB-FUTURE, showing the impact of each architectural component and the proposed loss function in \methodName.} 
  \vspace{-6pt}
  \label{tab-components}
  \centering
  \small
  \scalebox{0.95}{
  \begin{tabular} {lcccccc|cc} 
      \toprule 
      \textbf{Method} & RMSE$\downarrow$ & $RMSE_{log}\downarrow$  & Abs Rel$\downarrow$ & Sq Rel$\downarrow$  & log10$\downarrow$ & $\delta < 1.25\uparrow$ & OB-Recall$\uparrow$ & OB-Fscore$\uparrow$ \\
      \midrule

       Depth Baseline~\cite{yuan2022newcrfs} & 0.4524 & 0.1149 & 0.1011 & 0.0655 & 0.0327 & 0.9215 & - & - \\

       OB Baseline (ours) & - & - & - & - & - & - & 0.7655 & 0.5634 \\
       
       \midrule
       \textbf{S}hared encoder + \textbf{T}wo decoder (ST)  & 0.4376 & 0.1117 & 0.0983 & 0.0616 & 0.0414 & 0.9265 & 0.9154 & 0.6042  \\
       ST + \moduleName~& 0.4051 & 0.1038 & 0.0907 & 0.0536 & 0.0385 & 0.9415 & 0.9388 & 0.5621 \\
       ST + \moduleName~+ \lossName~& 0.3960 & 0.1007 & 0.0883 & 0.0503 & 0.0372 & 0.9494 & 0.9287 & 0.5895 \\
       
       \midrule
       Ours (ST + \moduleName~+ \lossName~+ \emph{EIP}) & 0.3963 & 0.1020 & 0.0901 & 0.0523 & 0.0380 & 0.9427 & 0.9090 & \textbf{0.6131} \\
       Ours (w/o \emph{EIP}) + \emph{SSR} & 0.3863 & 0.0983 & 0.0845 & 0.0469 & 0.0363 & 0.9516 & 0.9437 & 0.5510 \\
       Ours + \emph{SSR} & \textbf{0.3809} & \textbf{0.0974} & \textbf{0.0843} & \textbf{0.0468} & \textbf{0.0361} & \textbf{0.9518} & \textbf{0.9486} & 0.5415 \\
       
      \bottomrule 
  \end{tabular}
  }
  \vspace{-9pt}
\end{table*}

Our comprehensive ablation study in Table~\ref{tab-AGT} evaluated the effectiveness of training depth estimation alongside various auxiliary tasks through \methodName~on the \syntheticBenchmarkName. Beyond OBs, we compare three edge-based supervision variants: 1) semantic contours derived from segmentation masks, 2) instance boundaries from object instances, and 3) pseudo edges generated via LDC~\cite{soria2022ldc} (pretrained on BRIND~\cite{Pu_2021ICCV_brind}) with non-maximum suppression post-processing (aligned with standard practices in~\cite{he2019bdcn,feng2021mtorl}). Additionally, we benchmarked against surface normals and semantic segmentation-tasks conventionally correlated with depth.
Quantitative results reveal three key findings:
(1) Learning with auxiliary GTs improves depth results compared to the depth-only approach, validating the benefits of MT learning;
(2) Auxiliary OB supervision achieves the best overall performance across most metrics, particularly on RMSE and Sq Rel. In contrast, semantic segmentation supervision yields the best Abs Rel (with OB supervision second best) but performs relatively poorly on other depth metrics;
(3) Compared with other edge-type auxiliary GTs (second block in Table~\ref{tab-AGT}), OB supervision achieves the best performance on all depth metrics. While edge supervision shows competitive log10, its overall performance still lags behind OB supervision.
This systematic comparison demonstrates that OB supervision provides the most effective geometric constraints for MDE, leveraging the intrinsic relationship between depth discontinuities and OBs to improve both global accuracy and local geometric fidelity.

We conducted systematic component analysis in Table~\ref{tab-components} to validate \methodName's design, revealing four key findings:  
First, even the basic alternative model-consisting of Shared encoders and Two decoders-achieved significant performance improvements: RMSE improved from 0.4524 to 0.4376, while OB-Recall increased from 0.7655 to 0.9154. These demonstrate the value of joint estimation and highlight the inherent geometric reciprocity between two tasks.
Second, progressive integration of our components demonstrated hierarchical benefits-shared encoder features alone yielded 3.2\% RMSE for depth estimation and 7.1\% F-score improvements for OBs, while \moduleName~further improved 7.1\% RMSE through explicit boundary-depth feature modulation, crucially, when combined with \lossName, this configuration maintains depth estimation precision while optimally balancing \moduleName's dual impact on both OB estimation and depth refinement. 
Third, regarding \emph{EIP}, we observe an improved OB F-score in Stage One but a slight drop in depth metrics. This is because EIP sharpens true positives and suppresses non-OB features via attention to low-level details; however, this process may also filter out edge-like cues useful for depth regression, even though they are not true OBs.
Finally, in \emph{SSR} (mainly composed of the \moduleName), we observe higher depth metrics and OB-Recall but a slight drop in OB F-score. \moduleName~leverages mid-level OB features to boost depth prediction, as the depth stream learns to attend to a broader spectrum of edge information, but by enriching the depth feature space, it introduces a wider set of edge cues into the OB decoder, lowering OB F-score due to an increase in false positives.

%% file: sections_round2/5_conclusion.tex
\vspace{-4pt}
\section{Conclusion}
\label{sec:conclusion}
\vspace{-4pt}

In this work, we propose \methodName, a novel framework that pioneers joint depth and OB estimation, demonstrating OB's superior potential for modelling complex 3D scenes compared to alternative representations like edges or semantic contours. \methodName~achieves state-of-the-art performance on both synthetic benchmarks and the NYUD-v2 real-world dataset. Zero-shot cross-domain evaluations on iBims-1 and DIODE—without domain adaptation—further demonstrated our method's strong generalization capability in producing accurate depth maps with sharp OB discontinuities. Two critical limitations warrant acknowledgment. First, the absence of publicly available real-world datasets with paired depth and OB ground truth—a fundamental constraint in this emerging field—necessitated reliance on synthetic training data. This synthetic-real domain gap, particularly in OB representation fidelity, likely restricts our model's ability to handle real-world occlusion patterns. While our approach outperforms multi-task baselines, this limitation suggests significant room for improvement. Second, depth constraints receive relatively insufficient consideration within \moduleName, limiting OB prediction performance (\eg, F-score). Future work should address these limitations by: 1) developing techniques to improve cross-domain OB generalization, and 2) enhancing depth–OB interaction mechanisms within feature fusion modules.

\clearpage

\section*{Acknowledgment}

The authors would like to acknowledge Dr. Huan Fu, Dr. Xuchong Qiu and Prof. Mingming Gong for their valuable suggestions and assistance throughout this study. Furthermore, this work was supported by China Scholarship Council (CSC) scholarships awarded to Mr. Lintao Xu and Mr. Yinghao Wang.